\documentclass[11pt,a4paper]{article}
\usepackage[hyperref]{acl2021}
\usepackage{times}
\usepackage{latexsym}

\usepackage{amsmath}
\usepackage{graphicx}
\usepackage{bm}
\usepackage{mathrsfs}
\usepackage{amsfonts}
\usepackage{amssymb}
\usepackage{enumitem}
\usepackage[normalem]{ulem}
\usepackage{soul}
\usepackage{hyperref}
\usepackage {xcolor}
\usepackage{booktabs}
\usepackage{caption}
\usepackage{subcaption}
\usepackage{comment}
\usepackage{microtype}

\usepackage{multirow}
\aclfinalcopy % Uncomment this line for the final submission
\def\aclpaperid{34} %  Enter the acl Paper ID here

\title{Generating Faithful Text From a Knowledge Graph \\ with Noisy Reference Text}

\author{Tahsina Hashem$^1$, Weiqing Wang$^1$, Derry Tanti Wijaya$^2$, \\ \textbf{Mohammed Eunus Ali}$^3$, \textbf{Yuan-Fang Li}$^1$  \\ 
  $^1$Department of Data Science \& AI, Monash University, Australia \\ $^2$Department of Data Science, Monash University, Indonesia \\ $^3$Department of CSE, Bangladesh University of Engineering and Technology, Bangladesh  \\
  \small{\texttt{\{tahsina.hashem, Teresa.Wang, derry.wijaya, yuanfang.li\}@monash.edu}}; \\ \small{\texttt{eunus@cse.buet.ac.bd}} \\
  }

\date{}

\begin{document}
\maketitle
\begin{abstract}
Knowledge Graph (KG)-to-Text generation aims at generating fluent natural-language text that accurately represents the information of a given knowledge graph. While significant progress has been made in this task by exploiting the power of pre-trained language models (PLMs) with appropriate graph structure-aware modules, existing models still fall short of generating faithful text, especially when the ground-truth natural-language text contains additional information that is not present in the graph. In this paper, we develop a KG-to-text generation model that can generate faithful natural-language text from a given graph, in the presence of noisy reference text. Our framework incorporates two core ideas: 
Firstly, we utilize contrastive learning to enhance the model's ability to differentiate between faithful and hallucinated information in the text, thereby encouraging the decoder to generate text that aligns with the input graph. Secondly, we empower the decoder to control the level of hallucination in the generated text by employing a controllable text generation technique. We evaluate our model's performance through the standard quantitative metrics as well as a ChatGPT-based quantitative and qualitative analysis. Our evaluation demonstrates the superior performance of our model over state-of-the-art KG-to-text models on faithfulness. 
\end{abstract}

\section{Introduction}

A knowledge graph (KG) is a structured representation of information as a network of interconnected real-world entities, and relationships. The task of KG-to-text generation has been proposed ~\cite{ribeiro-etal-2020-modeling, koncel2019text} to make this structured information more accessible to humans, aiming to generate fluent, informative, and faithful natural-language sentences that should describe the contents of an input KG. Recently, this task plays a significant role in a variety of applications such as knowledge-grounded dialogue generation~\cite{kgdialzhou2018commonsense,kgdialzhao2020knowledge}, story generation~\cite{storyguan2019story,storyji2020language}, event narration~\cite{eventnarrative}, and question-answering~\cite{QAagarwal2021knowledge, QAchen2023toward, QAsaxena2020improving}.

Significant progress has been made in the KG-to-text generation task by utilizing a set of Transformer-based~\cite{vaswani2017attention} pre-trained language models (PLMs) such as BART~\cite{lewis2019bart}, T5~\cite{T5raffel2020exploring} or GPT~\cite{GPTradford2019language} with appropriate graph structure-aware modules~\cite{ke2021jointgt,colas2022gap,graphmaskhan2022self}. 
% While reducing hallucination in natural language generation has been an active research area~\cite{haldefsurveyji2022,haldefmaynez2020faithfulness}, 
However, ensuring the faithfulness of KG-to-text generation, i.e. reducing hallucinations~\cite{haldefsurveyji2022,controlTokenwang2022improving,haldefraunak2021curious,tablerebuffel2022controlling}, is an under-explored problem, and existing KG-to-text models fall short of generating faithful text when the ground-truth text of the training dataset contains wrong or extra information that is not consistent with the input. 
%~\cite{haldefsurveyji2022,haldefmaynez2020faithfulness,haldefraunak2021curious}. 
% Researchers define this situation as an intrinsic hallucination and an extrinsic hallucination problem respectively~\cite{haldefsurveyji2022,haldefmaynez2020faithfulness}. 

\begin{figure*}[htb]
     \centering
     \subfloat[House knowledge graph]{\includegraphics[width=0.58\textwidth]{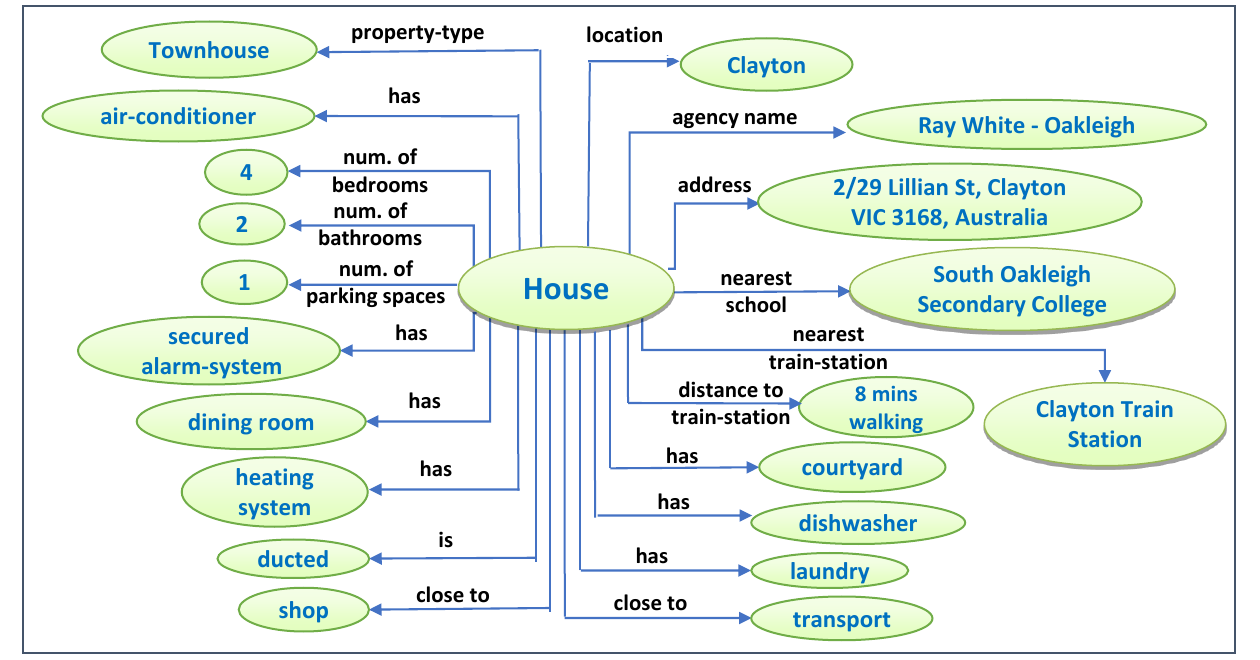}}\hfill
\subfloat[Ground-Truth Text]{\includegraphics[width=0.42\textwidth]{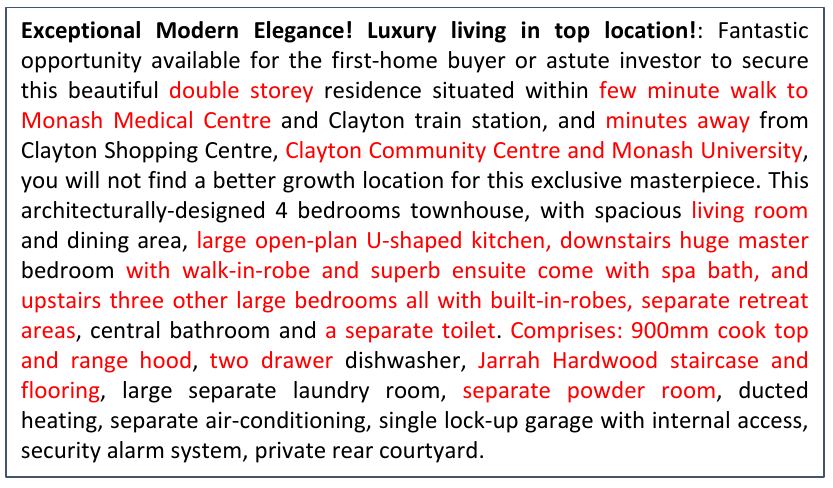}}
        \caption{A sample knowledge graph for the House dataset with its ground-truth text. The \textcolor{red}{red} colored text in the ground-truth text represents extrinsic hallucination information.}
        \label{fig:HouseKG}
        \vspace{-1em}
\end{figure*}

Figure~\ref{fig:HouseKG} shows an example of a small KG about a house, which contains information on its internal features and neighborhood, and the corresponding ground-truth reference text, from a real-world real-estate KG~\citep{das2021boosting}. The ground-truth text, while summarizing the features of the house accurately, also mentions some information that is not available in the input KG (i.e.\ extrinsic hallucination, highlighted in \textcolor{red}{red}).
%limits the performance of natural language generation models~\cite{confittang2021confit,cliffcao2021cliff,controlrashkin2021increasing,controlTokenwang2022improving}. 

When a KG-to-text model is trained with such hallucinated reference text, it is likely to produce text that is also hallucinated. This hallucination problem significantly reduces the faithfulness and thus trustworthiness of the generated text. Thus, the ability to reduce hallucination in the presence of noisy reference text is important for the practical application of KG-to-text and other NLG techniques, especially in mission- and safety-critical domains such as medical diagnostics and scientific research. % For example, a hallucinated summary of any scientific report or news gives users misleading information. Also in the medical field, a hallucinated patient’s medical test investigations may cause a serious negative impact on the patient.
%%%%%%%%%%%%%%%%%%%%%%%%%%%%%%%%%%%%%%%%%%%%%%%%%%%%%%%%%%%%%%%%%%%%%%%%%%%%%%%
%%%%%%%%%%%%%%%%%%%%%%%%%%%%%%%%%%%%%%%%%%%%%%%%%%%%%%%%%%%%%%%%%%%%%%%%%

A number of techniques have been proposed ~\citep{haldefsurveyji2022} to control this hallucination problem in abstractive summarization, table-to-text generation, generative question-answering, neural machine translation, and knowledge-grounded dialogue generation~\cite{controlTokenwang2022improving,confittang2021confit, tablerebuffel2022controlling, QAHalkrishna2021hurdles, TransHalDetectzhou2021detecting, controlzhang2022improving}. However, to the best of our knowledge, controlling hallucination in graph-to-text generation with noisy reference text has not been investigated. 

In this paper, we propose a novel framework to address this important and practical problem. 
Our framework combines contrastive learning technique and controllable text generation. %Researchers successfully applied these two techniques separately in controlling the hallucination problem in different natural language generation tasks~\citep{contrastDiageng2022improving,contrastdziri2022faithdial,contrastwan2022factpegasus,controlTokenwang2022improving,controlzhang2022improving,confittang2021confit,cliffcao2021cliff, filippova-2020-controlled, controlrashkin2021increasing}. 
Contrastive learning enables the model to distinguish between faithful and hallucinated text and guides the decoder to generate faithful text instead of hallucinated text. The controllable text generation technique learns the level of hallucination from noisy training text and controls (i.e.\ minimizes) the level of hallucinated information in the generated text. Our framework can be employed in any KG-to-Text encoder-decoder model to generate faithful natural language text from a given KG, in the presence of noisy reference text. %Experimental results demonstrate the efficiency of our proposed fine-tuning strategy on faithfulness metrics.

Our contributions are as follows:
\begin{itemize}[nosep]
  \item We propose a framework to deal with the hallucination problem in 
 KG-to-text generation task. Our framework comprises two core ideas: 
     (i) Employing contrastive learning to enable the KG-to-text generation model to better differentiate between faithful and hallucinated information in the reference text and guide the decoder to generate text that is faithful to KG. 
     (ii) Controlling the level of hallucination while generating text from KG using a controllable text generation technique.
     
  \item We conduct experiments and evaluate performance using a standard quantitative analysis with automatic metrics. Our comprehensive evaluation on two noisy datasets demonstrates the superior performance of our proposed model over the state-of-art KG-to-text generation models on faithfulness metrics. 

  \item We further propose and perform novel ChatGPT-based quantitative and qualitative evaluations to assess the performance of our model more comprehensively. The evaluation also shows our model's effectiveness in generating faithful text over existing KG-to-text generation models.  
\end{itemize}

\section{Related Work}
\label{sec:length}
\subsection{Knowledge Graph-to-Text Generation}
KG-to-text generation techniques~\citep{koncel2019text,guo2020cyclegt,ribeiro2020modeling,chen2020kgpt} utilize graph neural networks~\citep{velivckovicgraph} and graph Transformers~\citep{vaswani2017attention} to effectively encode a graph's structural information. With the rapid advancement of pre-trained language models (PLMs)~\citep{lewis2019bart,T5raffel2020exploring,GPTradford2019language}, researchers have started adapting and fine-tuning these models to KG-to-text generation tasks and obtained better results compared to previous models~\citep{ribeiro2021investigating,chen2020kgpt,kale2020text}. Recently, researchers further improved the KG-to-text models' performance by integrating pre-trained language models with appropriate graph-structure-aware modules~\citep{ke2021jointgt,colas2022gap} and employing some graph masking pre-training tasks~\citep{ke2021jointgt,graphmaskhan2022self}.%\eunus{can you be specific about it?}

However, we have empirically observed that although these state-of-art KG-to-text generation models~\citep {ke2021jointgt,colas2022gap,graphmaskhan2022self} introduce graph aware encoders and/or apply graph masking pre-training strategies to enhance graph-text alignments, still these models are struggling with hallucination problems when trained with noisy input ground-truth text.

\subsection{Controlling Hallucinations in Text Generation}
This hallucination problem is well explored in other natural language generation tasks such as in table-to-text generation, summarization, dialogue generation, question-answering, and neural machine translation. 
% In table-to-text generation tasks, hallucination mainly occurs due to the mismatch between the table and the reference text in the training data ~\citep{wikibiolebretneural,tottoparikh2020totto,WikipersonDatasun2022VNEL}.  
% An appropriate content plan or skeleton is prepared based on the input table and then the sequence generator is guided to generate text aligning with the given template employing reinforcement learning framework or edit-based generation technique. 
Planning ~\citep{tablePlansu2021plan} or skeleton-based method~\citep{tablesanawang2021sketch}, joint learning strategy~\citep{tableAggGenxu2021agggen}, Bayes training framework~\citep{tabConfidenttian2019sticking}, table-text optimal-transport matching strategy~\citep{tabwang2020towards}, control token approach~\citep{controlfilippova2020controlled} are widely used in controlling hallucinations in table-to-text generation tasks. Most recently, Rebuffel et al.~\cite{tablerebuffel2022controlling} proposed a multi-branch decoder approach to control hallucination at decoding time in this area. 

Prior works have also focused on minimizing hallucinations in summarization, dialogue generation, question-answering and neural machine translation areas. Some of the recent hallucination mitigation techniques are based on control token approach ~\citep{controlfilippova2020controlled,controlrashkin2021increasing,controlTokenwang2022improving}, contrastive learning approach~\citep{cliffcao2021cliff,confittang2021confit}, generate then-refine strategy~\citep{NeuralPAthdziri2021neural}, a routing transformer based approach~\citep{QAHalkrishna2021hurdles} and self-training of neural machine translation based approach~\citep{TransHalDetectzhou2021detecting}. To the best of our knowledge, no work has been done in graph-to-text generation tasks with hallucinated ground-truth text.
\subsection{Evaluation using ChatGPT}
% Large language models (LLMs) like GPT-3~\citep{GPT3brown2020language}, ChatGPT~\citep{ChatGPT}, GPT-4~\citep{GPT-4citekey} have achieved superior performance on a wide range of tasks, including natural language understanding (NLU)~\citep{NLUqin2023chatgpt,NLUbang2023multitask}, natural language generation (NLG)~\citep{NLGjiao2023chatgpt,NLGwang2023cross,NLGyang2023exploring}, and inference and reasoning~\citep{inferencezhong2023can,reasonerkojimalarge,reasonersweichain}. 
Large language models such as ChatGPT have recently been employed for evaluating the quality and factual consistency of the generated text in NLP tasks with respect to the source input through ranking, rating, and entailment inference~\citep{evaluatorkocmi2023large,evaluatorwang2023chatgpt,evaluatorluo2023chatgpt}. \citet{evaluatorluo2023chatgpt} closely investigated ChatGPT's ability under a zero-shot setting with three factual consistency evaluation tasks: binary entailment inference, summary ranking, and consistency rating. Experimental findings show that ChatGPT generally performs better than previous evaluation metrics across the three tasks, demonstrating its significant potential for factual consistency evaluation. However, they also point out some limitations of ChatGPT such as its preference on lexical similarity instead of semantic entailment, false reasoning, and poor understanding of instructions.
% Existing works measure the quality and factual consistency of the generated text in NLP tasks with respect to the source input through ranking, rating, and entailment inference~\cite{evaluatorwang2023chatgpt,evaluatorluo2023chatgpt,evaluatorkocmi2023large}. 
Moreover, while these approaches can compute an overall faithfulness score of the output text, they fall short in terms of explaining the score e.g., by quantifying the amount of hallucination (out of all the output facts, how many are hallucinated?), precision (out of all the output facts, how many are input facts?) and recall (out of all the input facts, how many appear in the output?). In this work, we use ChatGPT to quantify each of these values and obtain a finer-grained explanation of what a faithfulness score entails. 

\section{Proposed Model}
\subsection{Problem Formulation}
% In our research problem, given a set of KG-text pairs as a training dataset, the goal is to generate a faithful passage of text from a new KG. Formally, we can define our research problem as: 

Let $G$ = $(V, E)$ represent a knowledge graph, where $V=\{{e_1}, {e_2}, \dots, {e_{|V|}} \}$ represents the entity set and $E = \{r_{ij}\} \subseteq V\times V$ represents the relations connecting the entities, the task of KG-to-text aims to generate a passage of text $\hat{Y}$ = $( y_1, y_2, \ldots, y_n )$, that faithfully represents the information contained in a graph $G$. The model is given a training set $\mathcal{D} = \{(G_i, Y_i)\}$, in which the reference text $Y_i$ may contain \emph{hallucinated} information.

\subsection{Our Framework}
% Our main objective is to train a model to identify the faithful information from the hallucinated ground-truth text during the training \tw{maybe ``training'' instead of ``fine-tuning'' is better} process and develop the ability to minimize the occurrence of such hallucinated information while generating text from a KG. 
Standard fine-tuning approaches use a cross-entropy loss to maximize the similarity between the ground-truth text and the output text. Thus, if the ground-truth text contains hallucination, the model trained through fine-tuning also learns to generate hallucinated text. To overcome this hallucination problem, we introduce an effective fine-tuning approach that combines a contrastive loss function and a controllable text generation technique with the cross-entropy loss function. As a result, our method can train a KG-to-text generation model to generate faithful text from a KG.
%in order to generate a faithful and informative text from the knowledge graph. 

\begin{figure}[hbtp]
\begin{center}
\includegraphics[width=0.48\textwidth]{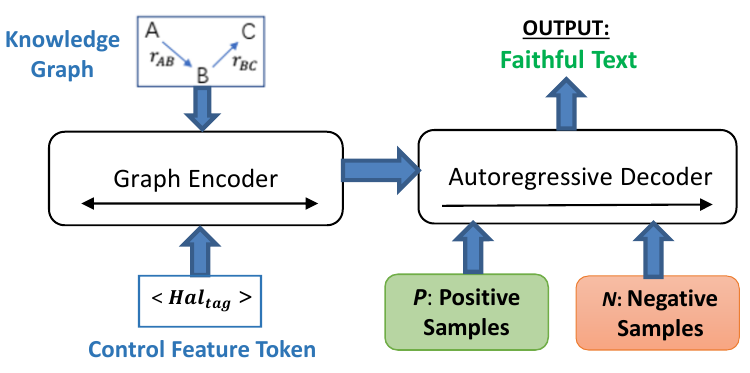}
\end{center}
\caption{The overall framework of our KG-to-text model.}
\label{fig:fullModel}
\vspace{-1.5em}
\end{figure}

Figure~\ref{fig:fullModel} depicts the overall architecture of our proposed model. The following two subsections illustrate our two proposed techniques in detail. 

% \tw{For this figure, 1) why there are two positive samples but only one negative samples? please refer to my comments under equation 1 2) it is suggested to have an overall introduction to the overall framework in 3.2 before we move the details in each component starting from 3.3; 3) My understanding is that the control token is applied to decoder while contrastive is applied to encoder according to our abstract. however, in this figure, it is the other way around}.
% \eunus{yes, agree Teresa as it contradicts with the abstract, you may swap the two in the figure?}
%%%%%%%%%%%%%%%%%%%%%%%%%%%%%% FIGURE 5%%%%%%%%%%%%%%%%%%%%%%

\subsection {Minimizing Hallucinations with Contrastive Learning}
Contrastive learning is a popular and effective representation learning method. Originally proposed for computer vision tasks ~\citep{khosla2020supervised, vision}, contrastive learning has been successfully applied to learn representations for sentences/documents ~\cite{gao2021simcse,zhang2021supporting}, abstractive summarization ~\cite{contrastSumliu2021simcls, cliffcao2021cliff, contrastwan2022factpegasus} and dialogue generation~\citep{confittang2021confit,contrastdziri2022faithdial,contrastDiageng2022improving}. Inspired by them, we have utilized this learning framework to reduce hallucinations while generating text from knowledge graphs. It enables the model to differentiate between faithful information and hallucinated information in the text, which then assists the decoder in generating text that should be free of hallucinations. 

% The contrastive loss function \cite{gunel2020supervised}, which we have used for fine-tuning is demonstrated in Equation~\ref{eq3.1}. 
For an input pair of a graph and an anchor reference text $(G_i, Y_i)$ from the training data $\mathcal{D}$, $P_i$ represents a set of positive samples and $N_i$ represents a set of hallucinated summaries (i.e.\ negative samples). The contrastive learning objective function is formulated as follows in Equation~\ref{eq3.1}:

\begin{equation} \label{eq3.1}
L_{CL} = - \sum_{(G_i, Y_i)\in\mathcal{D}} \sum\limits_{Y_j \in P_i} \log\frac{\exp(\cos(h_i,h_j))}{\sum\limits_{Y_k \in N_i}\exp(\cos(h_i,h_k))}
\end{equation}
Here, $Y_j$ is a positive sample from the set $P_i$, $Y_k$ is a negative sample from the set $N_i$, and $h_i$, $h_j$, $h_k$ are the BART decoder representations of $Y_i$, $Y_j$, and $Y_k$ respectively. %\tw{I think it is better in this way: i is the reference sample, j is the positive sample and k is the negative one}.
% \yf{I revised the above substantially, please have a look}

This contrastive objective function encourages the model to learn a preference for positive (faithful) summaries over negative (hallucinated) ones. 
While the ground-truth text in the training data $\mathcal{D}$ is noisy, it is reasonable to assume that each reference text is more faithful to the paired graph than a randomly sampled text from $\mathcal{D}$. Based on this observation, we  carefully select the positive and negative samples to ensure the effectiveness of our contrastive learning technique.

\textbf{Positive sample construction.} Back-translation ~\cite{backTransmallinson2017paraphrasing} is an effective approach for preserving meanings and providing linguistic diversity. Hence, we use NLPAug~\citep{NLPAugma2019nlpaug} to translate each anchor text to German and back to English and take the translated text as a positive sample for the anchor passage. % We use the German language here since it is a high-resource language and NLPAug is a good machine translation model. %Then we add the ground-truth text and it's best translation text to the positive samples set $P$.

\textbf{Negative sample construction.}
For the anchor text of a given graph, we treat the text of any other graph in $\mathcal{D}$ as a potential negative sample. We randomly select four such text to construct $N$ for each anchor text. Dataset-specific knowledge can be easily incorporated in this approach to improve the quality of contrastive learning. For the House dataset, we adopt a simple heuristic for constructing the negative sample set. Here, we give more importance to the six major features of a house graph: (1) house location (2) house address (3) number of bedrooms (4) number of bathrooms (5) number of parking spaces, and (6) house property type. If all of these major features of a house differ from the anchor house, then the house's paired text is selected as the negative sample for the anchor house. We choose these six features as major features because information of these features is available in almost every house (91\%)  in the training set.%, while the houses' internal and neighborhood information varies.
 
%In contrastive learning, the model is able to distinguish between faithful information and hallucinated information in the text. 
\subsection {Controlling Hallucinations with Control Feature Token}
In contrastive learning, we use the ground-truth reference text as a positive sample. As the ground-truth text contains hallucinations, when training with contrastive learning for generating text, the output text still contains some hallucinations.
Thus, we employ a controllable text generation approach to further enhance the faithfulness of our model. %~\cite{CTRLkeskar2019ctrl} for minimizing hallucinations in different natural language generation tasks but not in graph-to-text generation tasks. 
Specifically, we append controllable features to the input graph in training in order to control the level of hallucination in the generated text. 

\textbf{Control feature token.}
Control feature token is a hallucination measure that quantifies how much the given ground-truth text is faithful to the source graph. 
We linearized the knowledge graph~\citep{chen2020kgpt} into a list of verbalized triples and employ BARTScore~\cite{BARTScoreyuan2021bartscore} as the measure of faithfulness between the linearized graph and the corresponding ground-truth text, as it has been shown that it is closely associated with human evaluations of faithfulness~\cite{BARTScoreyuan2021bartscore}. 
% However, BARTscore is unable to input the KG directly. 
% We first linearized the knowledge graph~\citep{chen2020kgpt} into a list of verbalized triples. %, where each triple consists of the head entity, the relation/feature, and the tail entity sequentially. 

% BARTscore utilizes a pre-trained BART Transformer~\citep{lewis-etal-2020-bart} model that can score a linearized KG and the corresponding textual summary. It has been shown in \cite{BARTScoreyuan2021bartscore} that this metric is closely associated with human evaluations of faithfulness. Researchers applied this metric to measure the faithfulness of NLG models~\cite{confittang2021confit,bartscoregao2022dialsummeval}.

%%%%%%%%%%%%%%%%%%%%%%%%%%%%%% FIGURE 5%%%%%%%%%%%%%%%%%%%%%%
\begin{figure}[hbtp]
\begin{center}
\includegraphics[width=0.50\textwidth]{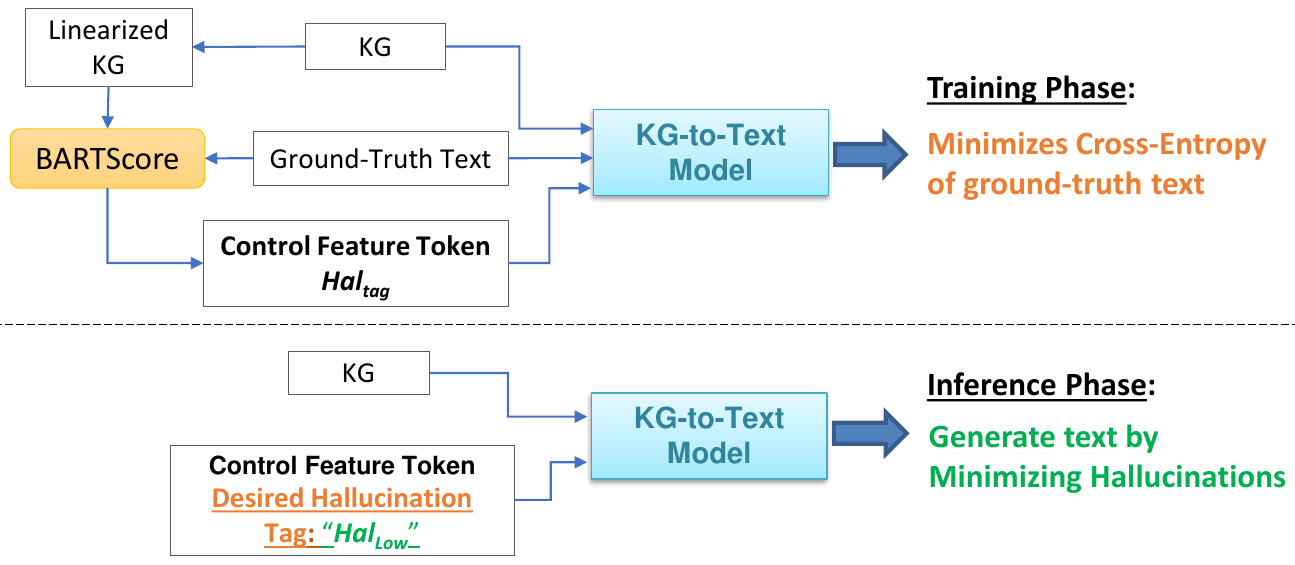}
\end{center}
\caption{Controllable Text generation with Control Feature Token}
\label{fig:controlTokenFig}
\end{figure}

% \begin{table*}[ht]
% \centering
% \small
% \begin{tabular}{ c|c|c|c|c  }
% \hline
%  \textbf{Dataset} &\textbf{\#Relations} &\textbf{\#KG-Text Pairs} &\textbf{\#Avg Triples} &\textbf{Avg Length}\\
%  & & (Train / Valid / Test) & & \\
%  \hline
%  House   &$68$ &$33$K / $10$K / $10,219$   &$24.69$ &$134.727$\\
%  GenWiki\textsubscript{FINE}& $287$   &$757,152$ &$2.64$$\pm$$1.72$ &$26.05$$\pm$$10.99$\\
% GenWiki\textsubscript{FULL} &$290$ &$1,336,766$  &$1.95$$\pm$$1.42$ &$21.46$$\pm$$10.64$\\
%  \hline
% \end{tabular}
% \caption{\label{table-splitdata}
% Statistics of fine-tuning datasets, including the total number relations, the data split, the average number of triples, and the average length of texts.
% }
% \end{table*}

\textbf{Controllable generation.}
% For every sample KG-text pair: we measure hallucination using BARTScore.  BARTScore gives us a measure of how much the given ground truth text is faithful with respect to the input linearized graph. The more faithful the ground truth text is, the less hallucinated the text is. 
According to the BARTScore of the training samples, we split the samples into three buckets, where each bucket contains a list of training samples at a specific range of BARTScore. This range is chosen in a manner that ensures each bucket contains approximately an equal number of samples. These three buckets are represented using the following hallucination tags, $Hal_{tag}$=\{$Hal_{low}$, $Hal_{medium}$ and $Hal_{high}$\} following existing work~\cite{controlfilippova2020controlled,controlzhang2022improving}. At training time, we append the corresponding hallucination tag to the input sample according to its BARTScore. These three hallucination tags represent the three control feature tokens that act as a special input to control the level of hallucination during text generation.

Figure~\ref{fig:controlTokenFig} illustrates the fine-tuning process with the control tokens. Let $G$ and $Y$ = $( y_1, y_2, \ldots, y_n )$ be the input sample graph and its corresponding reference text, and $H$ be the hallucination tag (i.e.\ control feature token) for this input sample. %Our goal is to generate a faithful text sequence $\hat{Y}$ = $(y_1, y_2, \ldots, y_n )$. 
Formally, we define the objective function of our fine-tuning strategy with the control token as follows:
\begin{equation}\label{eq3.2}
        L_{CE\_CtrlTok} = - \sum_{i = 1}^n log P (y_i|y_{<i}, G, H)
\end{equation}

Thus, during training, the model learns the mapping between the graph-text pair ($G$, $Y$) and its corresponding control token $H$. The model then becomes an expert at evaluating samples according to the control token. At inference time, the control token is set to the desired hallucinated value i.e., low (\textbf{$Hal_{low}$}) to generate faithful text from the KG.

\textbf{The overall training objective} of our proposed model is the sum of the contrastive loss and the cross-entropy loss with the control token:
\begin{equation}\label{eq3.3}
        L = L_{CL} + L_{CE\_CtrlTok}
\end{equation}

Thus, during training, instead of blindly following the ground-truth text, the model %creates a better representation of the input samples and 
gives more focus on the faithful parts of the text instead of the hallucinated ones. Moreover, the decoder is encouraged to generate text by minimizing hallucinations through controlled measures. %Experimental results show the effectiveness of our proposed contrastive fine-tuning task.

\section{Experiments}
\subsection{Dataset}
We conduct experiments and evaluation on two KG-to-text generation datasets: the House dataset~\citep{das2021boosting} about real-estate house listing and the GenWiki dataset~\cite{jin2020genwiki}. 
In both datasets, the ground-truth text contains a significant amount of hallucinated information, making the task of generating faithful text especially challenging. Thus, these datasets are the most appropriate to evaluate the performance of our proposed model. Table~\ref{table-splitdata} shows the statistics of these two datasets in detail. Note that we use the ``\textsc{fine}'' version~\cite{jin2020genwiki} of GenWiki. 

\begin{table}[ht]
\centering
\small
\begin{tabular}{lll}
\toprule
 \textbf{Dataset} &\textbf{\#Relations} &\textbf{\#KG-Text Pairs} \\
 & & (Train / Valid / Test) \\
 \midrule
 House   &$68$ &$33$K / $10$K / $10,219$ \\
 GenWiki\textsubscript{FINE}& $287$   &$750$K / $7,152$ / $1,000$ \\
% GenWiki\textsubscript{FULL} &$290$ &$1,336,766$ \\
 \bottomrule
\end{tabular}
\caption{\label{table-splitdata}
Statistics of the datasets, including the total number relations and the data split}
\vspace{-1em}
\end{table}
% Non-parallel dataset means~\citep{colas2021eventnarrative, jin2020genwiki}, there is a large disconnect between the graphs and their corresponding ground-truth texts. Here,
%% Many entities and relations from the triples of the graph are not contained within the text and 

\paragraph{House.}
The dataset is prepared from the large real-estate and POI datasets of Melbourne, Australia~\citep{das2021boosting}. %The dataset is used in a recent research work~\citep{das2021boosting} for predicting house prices. 
It includes $53,220$ records of house sales transactions from $2013$ to $2015$. It consists of three types of point-of-interests (POIs), namely regions, schools, and train stations, along with their corresponding features. Every sample in the dataset includes a ground-truth advertisement text describing the features of the house. However, the given ground-truth text contains a significant level of hallucinated information.

\begin{table*}[]
\small
\centering % used for centering table
\begin{tabular}{cccccc}
\hline\hline \\[-0.8em]
\multicolumn{6}{c}{\textbf{House Dataset}} \\ \hline\hline\\ [-0.8em]
\multicolumn{1}{l}{\multirow{2}{*}{\begin{tabular}[l]{@{}c@{}}\textbf{Model}\end{tabular}}} & \multicolumn{3}{c}{Comparison with ground-truth text}                                 & \multicolumn{2}{l}{Comparison with linearized graph}        \\ \cmidrule(lr){2-4} \cmidrule{5-6} \\[-0.9em]
\multicolumn{1}{c}{} & \multicolumn{1}{c}{BLEU $\uparrow$} & \multicolumn{1}{c}{METEOR $\uparrow$} & \multicolumn{1}{c}{ROUGE-L $\uparrow$} & \multicolumn{1}{c}{BARTScore $\uparrow$} & \multicolumn{1}{c}{FactCC $\uparrow$} \\ \hline \\ [-0.8em]
\multicolumn{1}{l}{Ground-truth text (5K samples)}  & \multicolumn{1}{c}{\textbf{-}} & \multicolumn{1}{c}{-} & \multicolumn{1}{c}{\textbf{-}}  & \multicolumn{1}{c}{-4.564}  & 48.48 \\ 
\multicolumn{1}{l}{JointGT~\citep{ke2021jointgt}}  & \multicolumn{1}{c}{\textbf{3.61}} & \multicolumn{1}{c}{11.96} & \multicolumn{1}{c}{\textbf{18.62}}  & \multicolumn{1}{c}{-3.685}  & 49.53 \\ 
\multicolumn{1}{l}{GAP~\citep{colas2022gap}}  & \multicolumn{1}{c}{3.47}    & \multicolumn{1}{c}{\textbf{12.05}} & \multicolumn{1}{c}{18.16}  & \multicolumn{1}{c}{-3.666}  & 52.71  \\ 
\multicolumn{1}{l}{GMP~\citep{graphmaskhan2022self}}  & \multicolumn{1}{c}{3.09}    & \multicolumn{1}{c}{10.73}  & \multicolumn{1}{c}{16.23}       & \multicolumn{1}{c}{-3.941}   & 48.47   \\ \midrule %\\[-0.6em]
\multicolumn{1}{l}{\textbf{Our Full Model}}  & \multicolumn{1}{c}{2.54}    & \multicolumn{1}{c}{11.06}      & \multicolumn{1}{c}{16.86}       & \multicolumn{1}{c}{\textbf{-3.245}}         & \textbf{63.61}  \\%[-0.8em] \\ 
\multicolumn{1}{l}{\quad Control token only}  & \multicolumn{1}{c}{2.88}    & \multicolumn{1}{c}{11.2}      & \multicolumn{1}{c}{17.35}       & \multicolumn{1}{c}{-3.567}         & 52.97   \\
\multicolumn{1}{l}{\quad Contrastive learning only}  & \multicolumn{1}{c}{2.56}    & \multicolumn{1}{c}{11.04}      & \multicolumn{1}{c}{16.89}       & \multicolumn{1}{c}{-3.247}         & 63.04   \\ \hline\hline \\ [-0.8em]
\multicolumn{6}{c}{\textbf{GenWiki Dataset}} \\ \hline\hline\\ [-0.8em]
\multicolumn{1}{l}{\multirow{2}{*}{\begin{tabular}[c]{@{}c@{}}\textbf{Model}\end{tabular}}} & \multicolumn{3}{c}{Comparison with ground-truth text}                                 & \multicolumn{2}{l}{Comparison with linearized graph}        \\ \cmidrule(lr){2-4} \cmidrule{5-6} \\[-0.9em]
\multicolumn{1}{c}{} & \multicolumn{1}{c}{BLEU $\uparrow$} & \multicolumn{1}{c}{METEOR $\uparrow$} & \multicolumn{1}{c}{ROUGE-L $\uparrow$} & \multicolumn{1}{c}{BARTScore $\uparrow$} & \multicolumn{1}{c}{FactCC $\uparrow$} \\ \hline \\ [-0.8em]
\multicolumn{1}{l}{Ground-truth text (5K samples)}  & \multicolumn{1}{c}{\textbf{-}} & \multicolumn{1}{c}{-} & \multicolumn{1}{c}{\textbf{-}}  & \multicolumn{1}{c}{-3.464}  & 53.80\\ 
\multicolumn{1}{l}{CycleGT~\citep{guo2020cyclegt}}  & \multicolumn{1}{c}{\textbf{41.59}}    & \multicolumn{1}{c}{\textbf{35.72}}      & \multicolumn{1}{c}{\textbf{63.31}}       & \multicolumn{1}{c}{-3.276}         & 76.86      \\ 
\multicolumn{1}{l}{JointGT~\citep{ke2021jointgt}}                                                                & \multicolumn{1}{c}{37.93}    & \multicolumn{1}{c}{32.60}      & \multicolumn{1}{c}{59.06}       & \multicolumn{1}{c}{-2.299}         & 79.94                           \\ 
\multicolumn{1}{l}{GMP~\citep{graphmaskhan2022self}}                                                                & \multicolumn{1}{c}{35.43}    & \multicolumn{1}{c}{32.68}      & \multicolumn{1}{c}{57.63}       & \multicolumn{1}{c}{\textbf{-1.601}}         & 76.62                           \\ \midrule %\\[-0.6em]
\multicolumn{1}{l}{\textbf{Our Full Model}}                                                 & \multicolumn{1}{c}{37.48}    & \multicolumn{1}{c}{32.70}      & \multicolumn{1}{c}{60.40}       & \multicolumn{1}{c}{-2.182}         & \textbf{82.85}                           \\[-0.8em] \\
\multicolumn{1}{l}{\quad Control token only}  & \multicolumn{1}{c}{37.01}    & \multicolumn{1}{c}{32.38}      & \multicolumn{1}{c}{59.57}       & \multicolumn{1}{c}{-2.268}         & 81.98   \\
\multicolumn{1}{l}{\quad Contrastive learning only}  & \multicolumn{1}{c}{35.19}    & \multicolumn{1}{c}{31.33}      & \multicolumn{1}{c}{57.89}       & \multicolumn{1}{c}{-2.309}         & 81.48 \\ 
% \multicolumn{1}{l}{\quad Contrastive learning only}  & \multicolumn{1}{c}{35.30}    & \multicolumn{1}{c}{31.44}      & \multicolumn{1}{c}{57.86}       & \multicolumn{1}{c}{-2.263}         & 81.67 \\ 
\hline\hline
\end{tabular}
\caption{Results on the \textbf{House} and \textbf{GenWiki} datasets. We have used BART-base and T5-base for House dataset and Genwiki dataset respectively. \textbf{Bold} fonts denote the best results.}
\label{table:main_results}
% \yf{Need to say which base PLM is used for each experiment (i.e.\ T5 and BART)}
\vspace{-1em}
\end{table*}

\paragraph{GenWiki.} It is a large-scale non-parallel~\citep{colas2021eventnarrative} 
% \tw{what does ``non-parallel'' mean here?} 
dataset prepared by matching Wikipedia articles with DBpedia entities~\citep{jin2020genwiki}. 

% Genwiki offers two versions of the dataset: the complete GenWiki\textsubscript{FULL} ($1.3$M) and the refined version GenWiki\textsubscript{FINE} ($750$K), which has stronger entity alignment between the graph and the text.\eunus{not clear why we choose the FINE version? Can er just avoid mentioning the full dataset?}
%Parallel and Non-parallel Dataset

\subsection{Baseline Models}
We evaluate the performance of our proposed model against graph-to-text generation models that are based on an encoder-decoder architecture. On the House dataset, we choose three state-of-the-art models: JointGT model~\citep{ke2021jointgt} that jointly learns the graph structure and text; GAP~\citep{colas2022gap} that is aware of the graph structure; and GMP~\cite{graphmaskhan2022self}, a self-supervised graph masking pre-training model. On the GenWiki dataset, we compare the results of the following models: the state-of-the-art unsupervised model CycleGT~\citep{guo2020cyclegt} for Genwiki dataset, JointGT (T5) model~\citep{ke2021jointgt} and GMP~\cite{graphmaskhan2022self}. Note that in addition to the existing state-of-the art model, GMP, we also include CycleGT as it has the best reported performance on GenWiki dataset.
% \tw{I think we still need a sentence to justify why Gap is used for House data while CycleGT is used for Genwiki data}

% \yf{The following are not baselines, and can be moved to the ablation study section}

% \eunus{is it better/ok to avoid the name JointGT from our final model?}

\subsection{Experimental Settings}
We adopt JointGT~\citep{ke2021jointgt} as our base model for fine-tuning. JointGT is initialized with the Hugging Face's pre-trained BART-base checkpoint\footnote{\url{https://huggingface.co/facebook/bart-base}} for House Dataset. For GenWiki dataset the model is initialized with the Hugging Face's pre-trained T5-base checkpoint\footnote{\url{https://huggingface.co/t5-base}}. We select the pre-trained LM BART-base or T5-base in order to do a fair comparison with the baseline models. 
% \eunus{do we need to say why we use BART FOR house and T5 for Genwiki, any suitable eplanation} 

JointGT is pre-trained with a KGTEXT dataset ~\citep{chen2020kgpt}. For contrastive learning, we use two positive samples and four negative samples for each training sample. For the House dataset, we fine-tune our model for 5 epochs; for the GenWiki dataset, we fine-tune our model for 4000 steps. The batch size is set to $32$. The maximum length of linearized input graphs is 600 and the maximum length of text sequences is set to 128 tokens. We adopt Adam~\citep{Adam} as the optimizer and set the learning rate to be 3e-5. We used one A40 48GB GPU and one A10 24GB GPU for the experiments 

% \tw{is our code ready for publication? if it is ready, it is better to put the link here}.

\subsection{Main Results}
We use automatic metrics to measure both fluency and faithfulness of generated text. Following existing KG-to-text work, we employ standard metrics 
BLEU~\citep{BLUEpapineni2002bleu}, METEOR~\citep{METEORbanerjee2005meteor}, and ROUGE-L~\citep{ROUGElin2004rouge}. These metrics are usually used to measure accuracy and fluency of the generated text with respect to the ground-truth text. However, as the ground-truth text contains hallucinations, we cannot verify the faithfulness of the generated text by comparing with these metrics. Thus, we use BARTScore~\cite{BARTScoreyuan2021bartscore} and FactCC~\cite{FactCCkryscinski2020evaluating} for comparing the generated text with the linearized input graph for measuring faithfulness. These two metrics have been widely used for measuring faithfulness in other NLP tasks~\cite{confittang2021confit, bartscoregao2022dialsummeval,cliffcao2021cliff,van2022mutual}. 

The faithfulness of the reference text of the House dataset and the GenWiki dataset is also reported in Table~\ref{table:main_results}, as measured by BARTScore and FactCC score. As can be seen, the reference text of both datasets contains significant amounts of hallucination (low BARTScore and FactCC scores). 
 
Table~\ref{table:main_results} presents the results on the House and GenWiki datasets. %We compute BLEU, METEOR, and ROUGE-L scores to measure the quality of the output text with respect to the ground-truth text. BARTScore and FactCC scores are calculated to measure the faithfulness of the generated text with respect to the input linearized graph.
From the results on the House dataset, we can observe that our full model achieves best results on faithfulness measures (i.e.\ when compared with the linearized graph), outperforming the best baseline models on BARTScore and FactCC score by 0.421 and 10.9 absolute points respectively. The performance delta on the GenWiki dataset is smaller, where our model achieves the best performance on FactCC of 1.55 points and second best performance on BARTScore. We posit the larger performance delta on the House dataset is due to it being significantly more noisy evidenced by lower BARTScore and FactCC scores. 

For BLEU, METEOR and ROUGE-L, the baseline models perform modestly better than our model when comparing with the ground-truth text. This result is expected and reasonable as compared with our model, the other models tend to generate text with higher similarity with the ground-truth text, resulting in higher values as measured by these metrics. At the same time, due to the noisy nature of the reference text, a high similarity also indicates high hallucination, as discussed above. 

In Section~\ref{sec:chatgpt} below, we further measure the faithfulness and fluency of generated text with ChatGPT as the oracle, where we demonstrate that our model achieves superior faithfulness while maintaining fluency. 
% We observe similar results on GenWiki dataset.
%indicates low faithfulness, since the ground-truth text contains large amounts of hallucination. 
% \yf{Can we say something about the fluency of our results?}

Table~\ref{fig:house} shows a sample ground-truth text and the text generated by different models, where correct facts are highlighted in \textcolor{blue}{blue} and hallucinated text is highlighted in \textcolor{red}{red}. More examples can be found in Appendix~\ref{sec:samples}.

\begin{table*}[htb!]
\small
\centering % used for centering table
\begin{tabular}{|p{5.9in}|}% centered columns  (2 columns)
\hline %inserts double horizontal lines
% \\  [-0.5em]
% \small{\textbf{House Knowledge Graph:}}  \\  [0.2em]
% \hline
%   \begin{center}
%       \includegraphics[scale=0.50]{Figure/ExampleFigureNew.pdf}
%   \end{center}\\[-0.7em]
%  % inserts table %heading
% \hline
\\[-0.5em]
\textbf{\underline{Ground-truth text:}} \\\\[-0.8em] \textbf{Exceptional potential with elevated outlook:}Set on the high side in a wonderfully quiet and convenient cul-de-sac location, this clinker-brick home provides an ideal opportunity to \textcolor{red}{renovate and extend, rebuild or redevelop (STCA)} and capture the \textcolor{red}{city skyline views}. Immediately comfortable, the home currently offers \textcolor{blue}{five bedrooms} (\textcolor{red}{two} with \textcolor{blue}{built-in robes}) and central \textcolor{blue}{bathroom}, \textcolor{red}{two living areas and well-appointed kitchen}. Other features include \textcolor{blue}{heating and cooling, expansive backyard, plus garage and carport} via full-length side driveway. \textcolor{blue}{Close to Rosehill Road/Hoffmans Road buses, shops} and \textcolor{red}{cafes}, \textcolor{red}{walking distance to St Bernard???s College and Rosehill Secondary College}.  \textcolor{red}{Currently tenanted at \$1653.00pm}. \textcolor{blue}{ Land size 15.8m x 42.1m ??? 665sqm (approx).} This is the one!\\ [0.3em]
\hline
\\[-0.5em]
\textbf{\underline{JointGT model:}} \\\\[-0.8em] \textcolor{red}{\textbf{AUCTION THIS SATURDAY AT 11:30AM!}}:In one of \textcolor{blue}{Niddrie's} most sought-after pockets, this solid-brick home presents an exciting opportunity to renovate and extend (STCA) on a deep block of land \textcolor{red}{(approx 690sqm)}. Current floorplan includes \textcolor{blue}{five bedrooms (all with built-in robes)} and central \textcolor{blue}{bathroom}, \textcolor{red}{formal lounge} through to the \textcolor{red}{kitchen/meals area}, as well as a separate \textcolor{blue}{toilet}. Other features of the home include \textcolor{blue}{gas heating and split-system air conditioning}, plus a north-facing \textcolor{blue}{backyard} complete with \textcolor{blue}{double garage} (remote-controlled  \\  [0.3em]
\hline
\\[-0.5em]
\textbf{\underline{GAP model:}} \\\\[-0.8em]
\textbf{Spacious family home in a sought-after \textcolor{blue}{Niddrie location!}}:Set on a large block of \textcolor{red}{land (approx 697sqm)}, this solid-brick home presents an exciting opportunity to renovate, extend or redevelop (STCA). Current floorplan includes \textcolor{blue}{five bedrooms} (two with \textcolor{blue}{BIRs}) and central \textcolor{blue}{bathroom}, \textcolor{red}{open-plan living/meals area and updated kitchen with stainless steel appliances}. Other features include \textcolor{blue}{split-system heating/cooling}, \textcolor{red}{polished floorboards, polished hardwood floors} and a \textcolor{blue}{double garage} with internal access. \textcolor{blue}{Close to schools, shops and public transport}, it's a must-see\\  [0.3em]
\hline
\\[-0.5em]
\textbf{\underline{GMP model:}} \\\\[-0.8em]
\textbf{Exceptional family living in a prime} \textcolor{blue}{\textbf{Niddrie location}}\textbf{!}:In a quiet cul-de-sac \textcolor{blue}{close to schools, shops and transport}, this impressive home offers an immediately comfortable family lifestyle with plenty of scope \textcolor{red}{to further improve and extend (STCA)}. Beyond its attractive facade, the home's inviting interior comprises \textcolor{blue}{five bedrooms (all with built-in robes)} and stylish central \textcolor{blue}{bathroom}, \textcolor{red}{spacious lounge through to the kitchen/meals area and adjoining family room}. Other highlights include \textcolor{blue}{ducted heating and evaporative cooling}, \textcolor{red}{polished hardwood floors}, plus \textcolor{blue}{a double garage} (with\\  [0.3em]
\hline
\\[-0.5em]
\textbf{\underline{Our model:}} \\\\[-0.8em]
\textcolor{blue}{\textbf{SOLD by Frank Dowling Real Estate}}: Located in a quiet pocket of \textcolor{blue}{Niddrie}, this immaculately presented family home is sure to impress. Comprising of \textcolor{blue}{5} good sized \textcolor{blue}{bedrooms with BIRs}, master with WIR and walk in robe, central \textcolor{blue}{bathroom}, \textcolor{red}{open plan living and dining area}, \textcolor{red}{central kitchen with stainless steel appliances}, family bathroom and separate \textcolor{blue}{toilet}. Other features include \textcolor{blue}{gas heating, evaporative cooling, split system air conditioner}, \textcolor{blue}{double garage} with internal access and a large \textcolor{blue}{rear yard} with rear access. \textcolor{blue}{Close to schools, shops and transport}. \\ [0.3em]
\hline
\end{tabular}
\caption{An example of ground-truth and generated text on the House dataset. Here \textcolor{red}{red} colored text represents \textcolor{red}{hallucinated information} and \textcolor{blue}{blue} colored text represents the \textcolor{blue}{faithful information.}} % title of Table
\label{fig:house}
\vspace{-12pt}
\end{table*}

\begin{figure}[hbtp]
\begin{center}
\includegraphics[width=0.5\textwidth]
{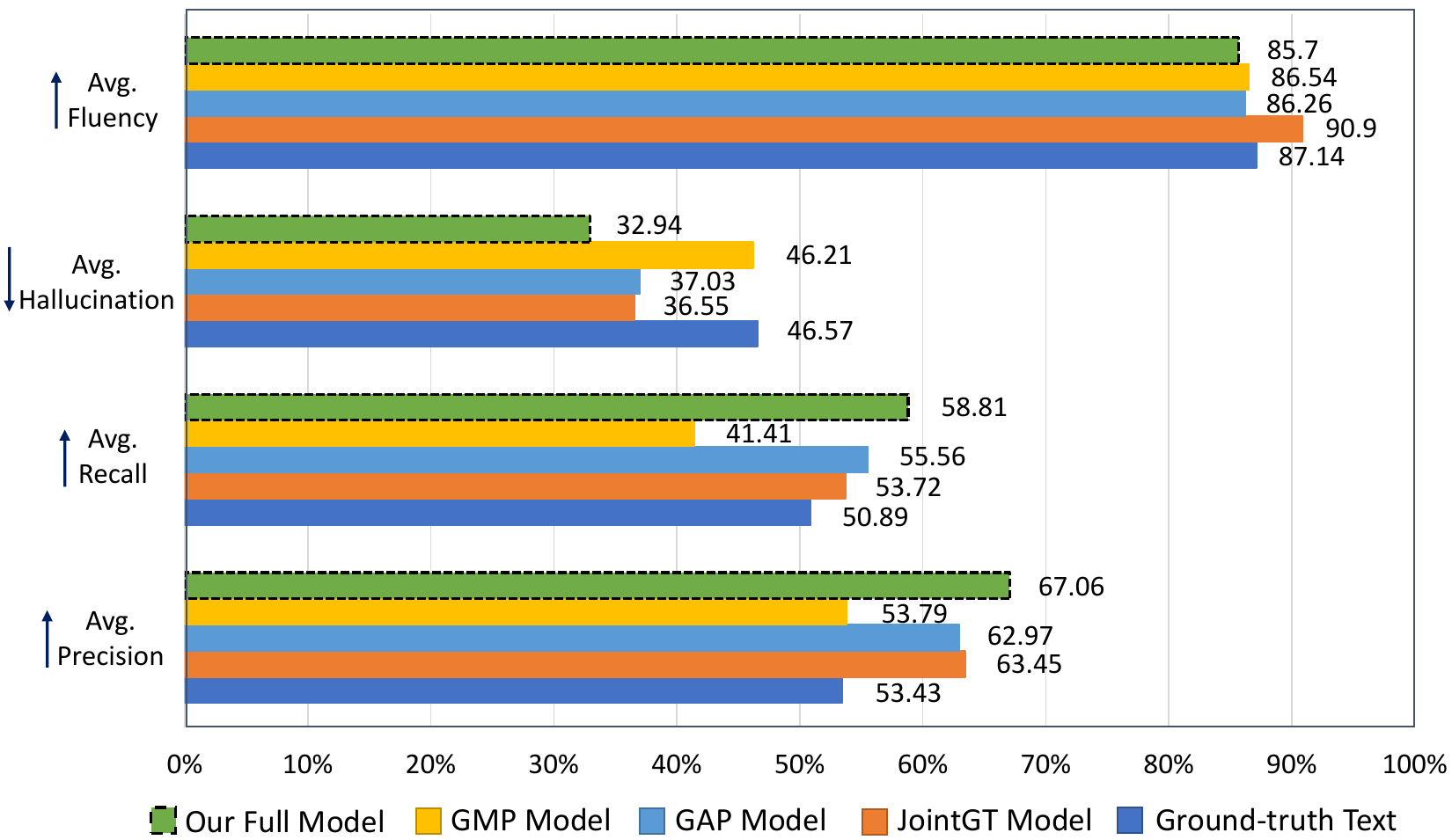}
\end{center}
\caption{ChatGPT-based evaluation on $50$ samples from the House test set.}
        \label{fig:chatGPTFigures}
\end{figure}

\subsection{ChatGPT-based Evaluation}\label{sec:chatgpt}
We propose to utilize ChatGPT to further measure the factual consistency and fluency of the generated text with respect to the input graph. 
We randomly sample 50 houses from the House test set, and perform evaluation on the text generated by different models. 

To measure \textbf{fluency}, similar to~\cite{evaluatorwang2023chatgpt}, we prompt ChatGPT to score the fluency of the generated text. 
%In this study, we conduct a detailed analysis to measure the faithfulness of the generated text according to the input graph using ChatGPT.
%For a better understanding of the graph, we input ChatGPT the linearized representation of the graph and its corresponding text generated by a KG-to-text model. 
To measure \textbf{factual consistency}, we carefully design prompts to instruct ChatGPT to enumerate facts in the (linearized) graph (\textit{\small\# input facts}), the common facts between the graph and generated text (\textit{\small\# common facts}), and the hallucinated facts in the generated text (\textit{\small\# hallucinated facts}), respectively. By enumerating facts that are correctly generated, missing, or hallucinated, our ChatGPT-based evaluation provides better explainability of models' faithfulness.  
% The prompt templates are shown in Figure~\ref {fig:PromptChatGPTTemplate}. %shows the ChatGPT prompt template that we have used to list the facts from the input graph and the model-generated text. 
Details and examples of our prompts and ChatGPT's responses can be found in Appendix \ref{app-prompt-design}. 

In addition to enumerating the facts, ChatGPT-based evaluation provides a way to measure quantitative metrics such as precision, recall, and hallucination rates. We randomly sample $50$ graph-text pairs from the test House dataset, and measure the precision (P), recall (R) and amount of hallucination (H) in generated text of these samples, which are formulated as follows: $P = \frac{\text{\emph{ \# common facts}}}{\text{\emph{\# output facts}}}$, $R = \frac{\text{\emph{\# common facts}}}{\text{\emph{\# input facts}}}$, and $H~=~\frac{\text{\emph{\# hallucinated facts}}}{\text{\emph{\# output facts}}}$.

The number of output facts (\textit{\small\# output facts}) is computed by summing up the number of hallucinated facts (\textit{\small\# hallucinated facts}) and the number of common facts (\textit{\small\# common facts}). 
% \yf{Why not just count the number of facts in the output? Seems simpler} 

% \subsection{ChatGPT Evaluation Results}
Figure~\ref{fig:chatGPTFigures} shows the results of this analysis. It can be seen that our model outperforms all the baseline KG-to-text generation models on precision, recall and faithfulness (i.e.\ low hallucination) and achieves competitive scores in terms of fluency. 

To determine the gap between our model and the most capable language models, we also compare our model with ChatGPT on a set of 1,000 random samples from the House dataset in different settings. A comprehensive analysis of this experiment is presented in Appendix~\ref{comparison-with-chatgpt}. As can be expected, ChatGPT achieves significantly better performance in faithfulness in zero-shot setting. However, when given noisy ground-truth text as few-shot examples, ChatGPT generates hallucinated text similar to the ground-truth text, showing that it is also prone to noise in the reference text. Our model outperforms ChatGPT in this (3-shot) setting in terms of precision and hallucination (i.e., lower hallucination). 

\subsection{Ablation Studies}
To investigate the effect of contrastive learning and control token techniques individually, we experiment on both datasets with two configurations of our full model: one with control token only and the other one with contrastive learning only. 

% (i) \textbf{KG-to-Text generation Model with Control Token }: Here, we fine-tune the JointGT model using only Control Token approach.

% (ii) \textbf{KG-to-Text generation Model with Contrastive Learning }: Here, we  fine-tune the JointGT model using Contrastive learning without including Control Token.

% (iii) \textbf{KG-to-Text generation Model with Control Token and Contrastive Learning}: This is our proposed model with full architecture, here we fine-tune the baseline JointGT model using contrastive learning and Control Token approach.

As we see in Table~\ref{table:main_results}, both model components contribute to our model's better faithfulness, with contrastive learning making a larger impact in House dataset. % better than all the existing generation models in terms of faithfulness. 

% Thus, these experimental findings confirm the effectiveness of contrastive learning and controllable text generation technique in fine-tuning in preventing hallucinations in the generated text.
% \yf{This above paragraph can be removed if we need space}

\section{Conclusion}
In this paper, we have proposed a novel approach to generate faithful text from a knowledge graph having noisy ground-truth text. To ensure faithful text generation, we have introduced two key ideas: (i) contrastive learning to better differentiate between faithful and hallucinated information, (ii) control token to regulate the level of hallucination in the generated text. Experimental results on two noisy KG-to-text datasets demonstrates that KG-to-text model with our framework outperforms all the baseline models in terms of faithfulness metrics. Moreover, we have proposed a novel ChatGPT based evaluation technique for an in-depth quantitative and qualitative analysis, which further verifies the superior performance of our model on precision, recall and faithfulness. 

\noindent\textbf{Limitation and Future work} 
We have applied our proposed framework only in PLM based KG-to-text encoder-decoder model. In future, we plan to explore the hallucination problem in AMR (Abstract Meaning Representations) graph datasets, which can also preserve a number of meaningful semantic relations and widely used in NLP areas.

\section*{Ethical Considerations}
Our model utilizes existing pre-trained language model based KG-to-text generation model, thus the ethical concerns associated with these models would also be applicable to our proposed framework.

\section*{Acknowledgments}
This material is based on research sponsored by Defense Advanced Research Projects Agency (DARPA) under agreement number HR0011-22-2-0047. The U.S. Government is authorised to reproduce and distribute reprints for Governmental purposes notwithstanding any copyright notation thereon. The views and conclusions contained herein are those of the authors and should not be interpreted as necessarily representing the official policies or endorsements, either expressed or implied, of DARPA or the U.S. Government.
\bibliographystyle{acl_natbib}
\bibliography{anthology,acl2021}

% \appendix{\textbf{Appendix}}
%\section{Data Statistics}\label{data-stat}

% \newpage
\clearpage
\appendix
\section{Prompt Design for ChatGPT-based Evaluation}
\label{app-prompt-design}
\begin{figure*}[hbtp]
\vspace{-0.2cm}
\begin{center}
\includegraphics[width=0.95\textwidth]
{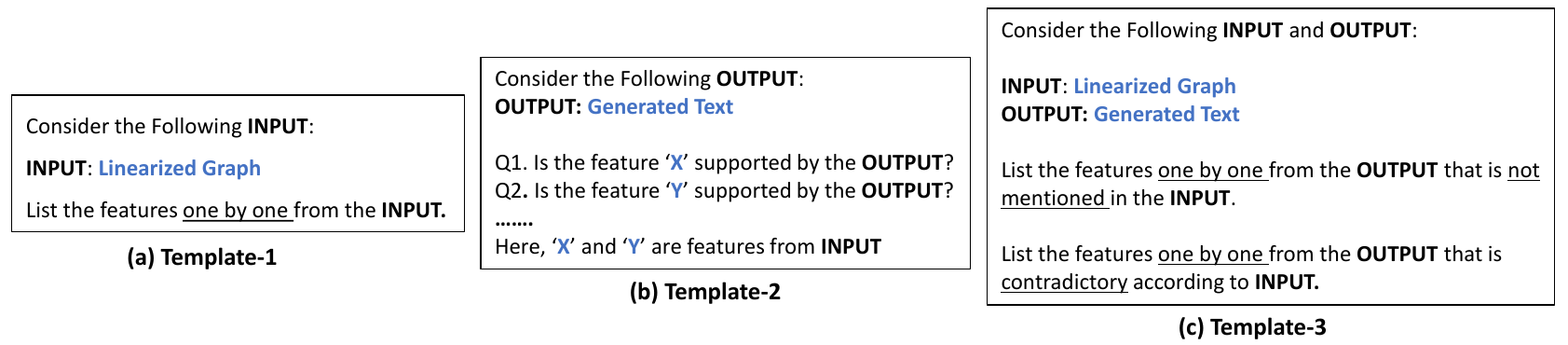}
\end{center}
\vspace{-0.2cm}
\caption{Prompt templates for enumerating facts using ChatGPT. Template-1 (left) is to enumerate facts in the input (linearized graph). Template-2 (middle) is to enumerate common facts between the input (linearized graph) and the output (generated text). Template-3 (right) is to enumerate hallucinated facts in the output (generated text).} 
% % \eunus{you may name these as Template 1, Template 2, Template 3, and the in subsequent examples, you can say An Example of Template 1 or An Example of Template 2...}}

\label{fig:PromptChatGPTTemplate}
\end{figure*}

The prompt templates are shown in Figure~\ref {fig:PromptChatGPTTemplate}.

\begin{figure*}[hbtp]
\begin{center}
\includegraphics[width=0.80\textwidth]
{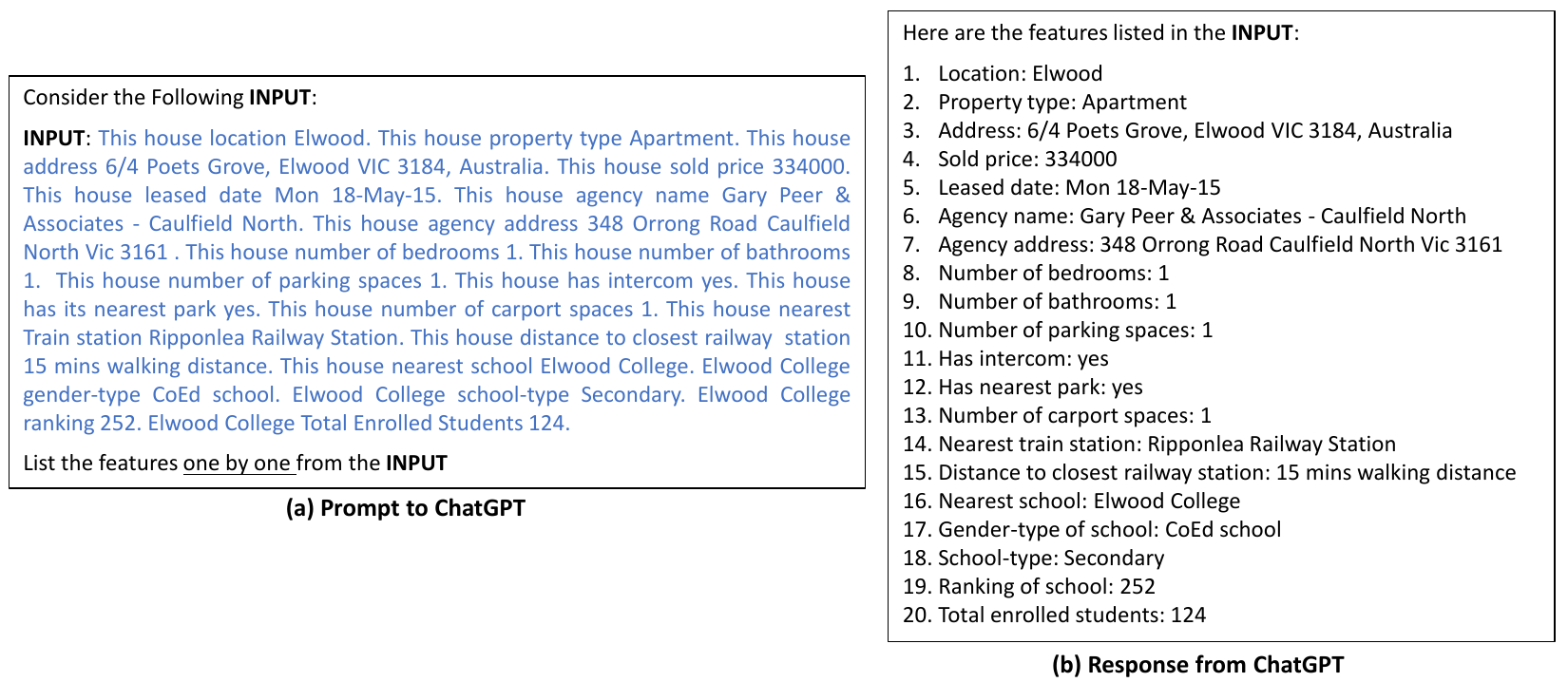}
\end{center}
\vspace{-0.2cm}
\caption{An Example of Template-1 }
\label{fig:InputFeatureschatGPT}
\end{figure*}

\paragraph{Listing the facts of a graph:}
Here we give ChatGPT an input linearized graph and ask it to ``list the features one by one from the INPUT" (Figure~\ref{fig:PromptChatGPTTemplate}-Left). Figure~\ref{fig:InputFeatureschatGPT} shows an example of this prompt to ChatGPT and its response for a sample from the House test set. ChatGPT has made no error in all $50$ test samples of House data.

\begin{figure*}[hbtp]
\begin{center}
\includegraphics[width=0.90\textwidth]
{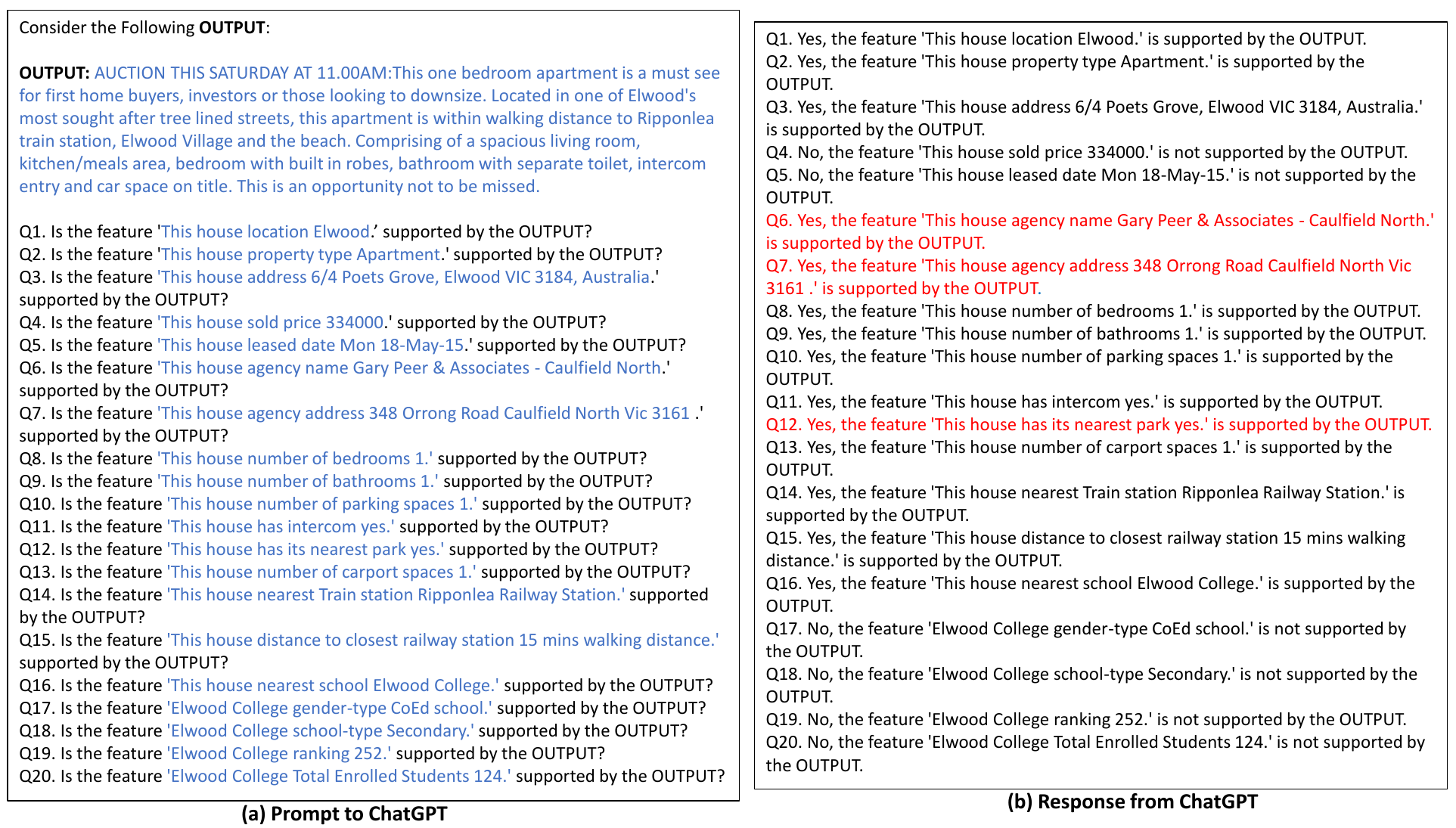}
\end{center}
\vspace{-0.2cm}
\caption{An Example of Template-2 }
\label{fig:MutualFeatureschatGPT}
\end{figure*}

\paragraph{Listing the common facts:}
ChatGPT was unable to correctly list the common facts between the linearized input graph and the generated text. Hence, we prompt ChatGPT for each fact listed in the input, whether that fact is included in the output. Here, each fact (or ``feature") represents a single triple of the input linearized graph (Figure~\ref{fig:PromptChatGPTTemplate}-Middle). Then, we count the answer with a ``yes'' response from ChatGPT. %The performance of ChatGPT is also good here. 
On average, ChatGPT makes 2-3 mistakes per sample. Figure~\ref{fig:MutualFeatureschatGPT} shows an example of this prompt and and ChatGPT's response. The red colored text indicates the mistakes done by ChatGPT.

\begin{figure*}[hbtp]
\begin{center}
\includegraphics[width=0.90\textwidth]
{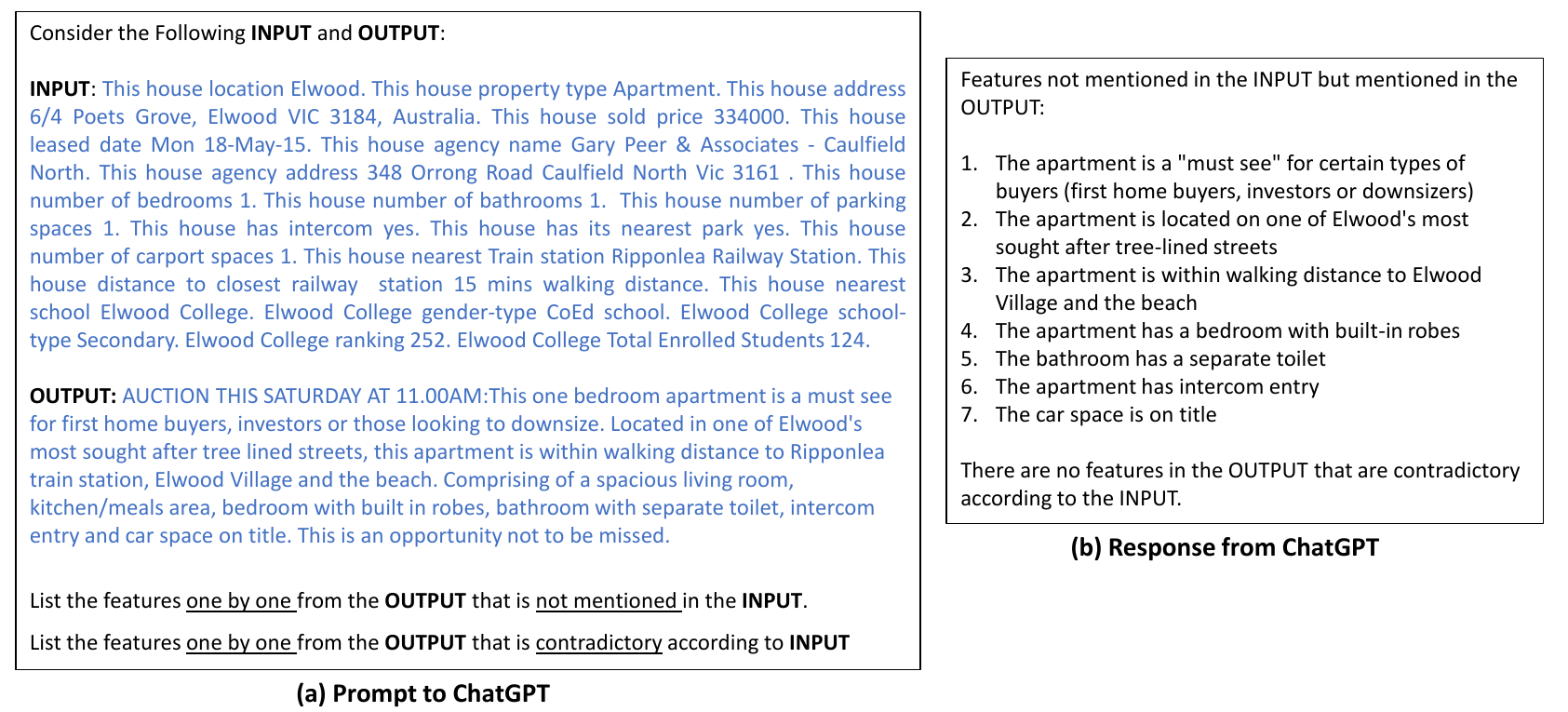}
\end{center}
\caption{An Example of Template-3 }
\label{fig:HallucinationFeatureschatGPT}
\end{figure*}

\paragraph{Listing the hallucinated facts:}
Here, we prompt ChatGPT to list both the extrinsic and intrinsic hallucination facts in the generated text by providing ChatGPT with an input (linearized graph) and an output (generated text). Firstly, to list the extrinsic hallucination facts we instruct ChatGPT to ``List the features one by one from the OUTPUT that is not mentioned in the INPUT''. Secondly, to list the intrinsic hallucination facts we instruct ChatGPT to ``List the features one by one from the OUTPUT that is contradictory to the INPUT'' (Figure~\ref{fig:PromptChatGPTTemplate}-Right). Here, ChatGPT makes no mistakes in the $50$ House test samples. Figure~\ref{fig:HallucinationFeatureschatGPT} illustrates the steps with an example and ChatGPT's response.

\begin{comment}
\section{ChatGPT Evaluation Details}
Figure~\ref {fig:InputFeatureschatGPT}, Figure~\ref {fig:MutualFeatureschatGPT} and Figure~\ref {fig:HallucinationFeatureschatGPT} show the examples of  ChatGPT prompt templates that we have used to list the features of the input graph and the output text.    
\end{comment}

%%%%%%%%%%%%%%%%%%%%%%%%%%%%%% FIGURE 5%%%%%%%%%%%%%%%%%%%%%%
%%%%%%%%%%%%%%%%%%%%%%%%%%%%%%%%%%%%%%%%%%%%%%%%%

%%%%%%%%%%%%%%%%%%%%%%%%%%%%%% FIGURE 5%%%%%%%%%%%%%%%%%%%%%%
%%%%%%%%%%%%%%%%%%%%%%%%%%%%%%%%%%%%%%%%%%%%%%%%%

%%%%%%%%%%%%%%%%%%%%%%%%%%%%%% FIGURE 5%%%%%%%%%%%%%%%%%%%%%%
%%%%%%%%%%%%%%%%%%%%%%%%%%%%%%%%%%%%%%%%%%%%%%%%%

\begin{figure*}[htbp]
\small
\centering % used for centering table
\begin{tabular}{|p{5.9in}|}% centered columns  (2 columns)
\hline %inserts double horizontal lines
\\  [-0.5em]
\small{\textbf{House Knowledge Graph:}}  \\  [0.2em]
\hline
  \begin{center}
      \includegraphics[scale=0.52]{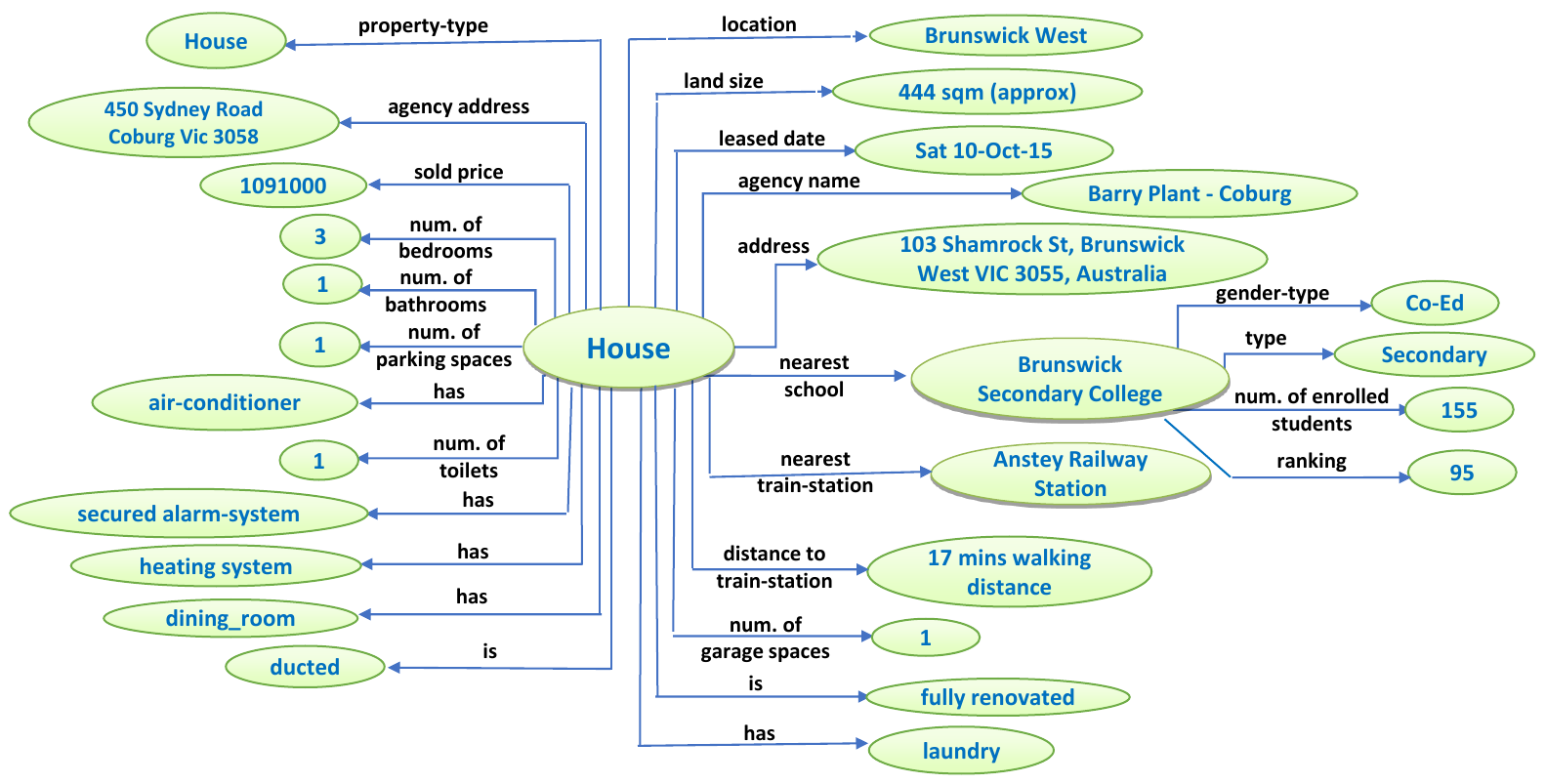}
  \end{center}\\[-0.7em]
 % inserts table %heading
\hline
\\[-0.5em]
\textbf{\underline{Ground-truth text:}} \\\\[-0.8em] \textbf{\textcolor{red}{AUCTION THIS SATURDAY 10TH OCTOBER AT 1PM:}} Superbly blending charm with a modern sensibility and fresh infusion of light, this fully renovated three bedroom \textcolor{red}{Californian Bungalow} offers stylish family living in a sought after locale.An expansive \textcolor{red}{open plan lounge} and dining room provides spacious living while showcasing \textcolor{red}{a marvellous kitchen boasting 700mm stainless steel appliances, abundant storage and feature island bench, ensuring effortless entertaining}.\textcolor{red}{A serene second living room, boasting leafy garden views} offers retreat from the bustle of family life.Three spacious bedrooms are serviced by a modern bathroom, which cleverly incorporates an \textcolor{red}{oversized European} laundry.Brilliantly positioned to transport, schools, cafes, vibrant Sydney Road, CityLink and featuring ducted heating, alarm, split system air-conditioning, BIR's, \textcolor{red}{ceiling fans} and carport \textcolor{red}{with OSP}, stylish family living starts here.\\ [0.3em]
\hline
\\[-0.5em]
\textbf{\underline{JointGT model:}} \\\\[-0.8em] \textbf{\textcolor{red}{AUCTION THIS SATURDAY AT 11:00AM:}} This \textcolor{blue}{renovated three bedroom} brick veneer home is sure to appeal to first home buyers, downsizers and investors alike.The home comprises of \textcolor{blue}{three} \textcolor{red}{double} \textcolor{blue}{bedrooms} (\textcolor{red}{all with built-in robes}), \textcolor{red}{spacious lounge} and \textcolor{blue}{dining room}, \textcolor{red}{renovated kitchen with stainless steel appliances}, separate \textcolor{blue}{laundry, ducted heating, evaporative cooling, alarm system} and a \textcolor{blue}{single garage} with internal access.The property also features \textcolor{red}{a large rear garden with established gardens and a bungalow}.  \\  [0.3em]
\hline
\\[-0.5em]
\textbf{\underline{GAP model:}} \\\\[-0.8em]
\textbf{\textcolor{blue}{Renovated} And Ready To Move In:} This beautifully \textcolor{blue}{renovated} brick veneer home is ready to move in and enjoy. Featuring a spacious \textcolor{blue}{living/dining area}, \textcolor{red}{renovated kitchen with stainless steel appliances and adjoining meals area}, \textcolor{blue}{three} \textcolor{red}{double} \textcolor{blue}{bedrooms} \textcolor{blue}{with built-in robes}, central \textcolor{blue}{bathroom and separate laundry}. Other features include \textcolor{blue}{ducted heating, evaporative cooling, alarm system, double garage with internal access}, \textcolor{red}{polished floorboards, polished hardwood floors, security shutters} and much more. \textcolor{blue}{Close to} \textcolor{red}{Sydney Road shops}, \textcolor{blue}{schools, public transport} and easy access to CBD.\\  [0.3em]
\hline
\\[-0.5em]
\textbf{\underline{GMP model:}} \\\\[-0.8em]
\textbf{\textcolor{red}{SOLD BY HARRY KONTOSSIS:}} \textcolor{blue}{Sale by SET DATE 10/10/2015} (unless sold prior) Situated in a quiet cul-de-sac in the \textcolor{blue}{heart of Brunswick West}, this immaculately presented \textcolor{blue}{three bedroom}, \textcolor{blue}{two bathroom} home is sure to appeal to first home buyers, investors and downsizers alike.Accommodation includes \textcolor{red}{a formal lounge} and \textcolor{blue}{dining room}, \textcolor{red}{well-appointed kitchen with stainless steel appliances, adjoining family/meals area}, \textcolor{blue}{three} \textcolor{red}{double} \textcolor{blue}{bedrooms} \textcolor{red}{with built-in robes}, central \textcolor{blue}{bathroom, separate toilet.}\\  [0.3em]
\hline
\\[-0.5em]
\textbf{\underline{Our framework:}} \\\\[-0.8em]
\textbf{\textcolor{blue}{Renovated} Family Home in the \textcolor{blue}{Heart of Brunswick West}:} This beautifully \textcolor{blue}{renovated} family home is set on a generous \textcolor{blue}{allotment of 444 sqm (approx)} and is ideal for first home buyers, downsizers or investors looking for a low maintenance lifestyle. Comprising of \textcolor{blue}{3 spacious bedrooms} (\textcolor{red}{master with WIR and walk in robe}), \textcolor{red}{formal lounge} and \textcolor{blue}{dining room}, \textcolor{red}{updated kitchen with stainless steel appliances}, central \textcolor{blue}{bathroom and laundry}. Other features include \textcolor{blue}{ducted heating, evaporative cooling, split system air conditioner, ducted cooling, alarm system, double garage} and much more. \\ [0.3em]
\hline
\end{tabular}
\caption{Example of generated text on the House dataset. Here \textcolor{red}{red} colored text represents \textcolor{red}{hallucinated information} and \textcolor{blue}{blue} colored text represents the \textcolor{blue}{faithful information}} % title of Table
\label{fig:houseExample}
\end{figure*}

\begin{table*}[htb]
\small
\centering % used for centering table
\begin{tabular}{cccccc}
\hline\hline\\ [-0.8em]
\multicolumn{1}{l}{\multirow{2}{*}{\begin{tabular}[l]{@{}c@{}}\textbf{Generation Model}\end{tabular}}} & \multicolumn{3}{c}{Comparison with ground-truth text}                                 & \multicolumn{2}{l}{Comparison with linearized graph}        \\ \cmidrule(lr){2-4} \cmidrule{5-6} \\[-0.9em]
\multicolumn{1}{c}{} & \multicolumn{1}{c}{BLEU $\uparrow$} & \multicolumn{1}{c}{METEOR $\uparrow$} & \multicolumn{1}{c}{ROUGE-L $\uparrow$} & \multicolumn{1}{c}{BARTScore $\uparrow$} & \multicolumn{1}{c}{FactCC $\uparrow$} \\ \hline \\ [-0.8em]
\multicolumn{1}{l}{ChatGPT-ZeroShot}  & \multicolumn{1}{c}{1.21} & \multicolumn{1}{c}{11.86} & \multicolumn{1}{c}{12.91}  & \multicolumn{1}{c}{\textbf{-2.389}}  & 71.02 \\ [0.2em]
\multicolumn{1}{l}{ChatGPT-1-Shot}  & \multicolumn{1}{c}{1.95}    & \multicolumn{1}{c}{12.73} & \multicolumn{1}{c}{15.02}  & \multicolumn{1}{c}{-2.872}  & \textbf{76.34}  \\ [0.2em]
\multicolumn{1}{l}{ChatGPT-2-Shot}  & \multicolumn{1}{c}{2.06}    & \multicolumn{1}{c}{12.67} & \multicolumn{1}{c}{15.58}  & \multicolumn{1}{c}{-2.937}  & 72.02  \\[0.2em]
\multicolumn{1}{l}{ChatGPT-3-Shot}  & \multicolumn{1}{c}{2.25}    & \multicolumn{1}{c}{\textbf{13.31}}  & \multicolumn{1}{c}{15.76}       & \multicolumn{1}{c}{-3.036}   & 73.88   \\ \midrule 
\multicolumn{1}{l}{\textbf{Our Full Model}}  & \multicolumn{1}{c}{\textbf{2.68}}    & \multicolumn{1}{c}{11.21}      & \multicolumn{1}{c}{\textbf{17.10}}       & \multicolumn{1}{c}{-3.246}         & 62.84  \\ \hline\hline \\ [-0.8em]
\end{tabular}
\caption{Results on $1000$ test samples from the House dataset. \textbf{Bold} fonts denote the best results.}
\label{table:chatgpt_results}
% \yf{Need to say which base PLM is used for each experiment (i.e.\ T5 and BART)}
\end{table*}

\begin{table}[htbp]
\small
\centering % used for centering table
\begin{tabular}{lccc}% centered columns  (2 columns)
\toprule %inserts double horizontal lines
\\[-1em]
\textbf{Generation }  & Avg. & Avg. & Avg. \\
\textbf{Model}  & Precision & Recall & Hallucination \\
 % inserts table %heading
\midrule
& & & \\[-0.7em]
ChatGPT-ZeroShot & \textbf{73.28} & \textbf{88.21} & \textbf{26.71} \\  [0.2em]
ChatGPT-3-Shot & 65.45 & 64.39 & 34.55 \\ 
\midrule
& & &\\[-0.7em]
\textbf{Our Full Model} & 67.06 & 58.81 & 32.94\\ [0.2em]
\bottomrule
\end{tabular}
\caption{ChatGPT Evaluation Results based on $50$ samples from the House Dataset. \textbf{Bold} fonts denote the best results.} % title of Table
\label{table:chatgptgenerations}
\end{table}

\begin{figure*}[htbp]
\small
\centering % used for centering table
\begin{tabular}{|p{5.9in}|}% centered columns  (2 columns)
\hline %inserts double horizontal lines
\\  [-0.5em]
\small{\textbf{Genwiki Knowledge Graph:}}  \\  [0.2em]
\hline
  \begin{center}
      \includegraphics[scale=0.50]{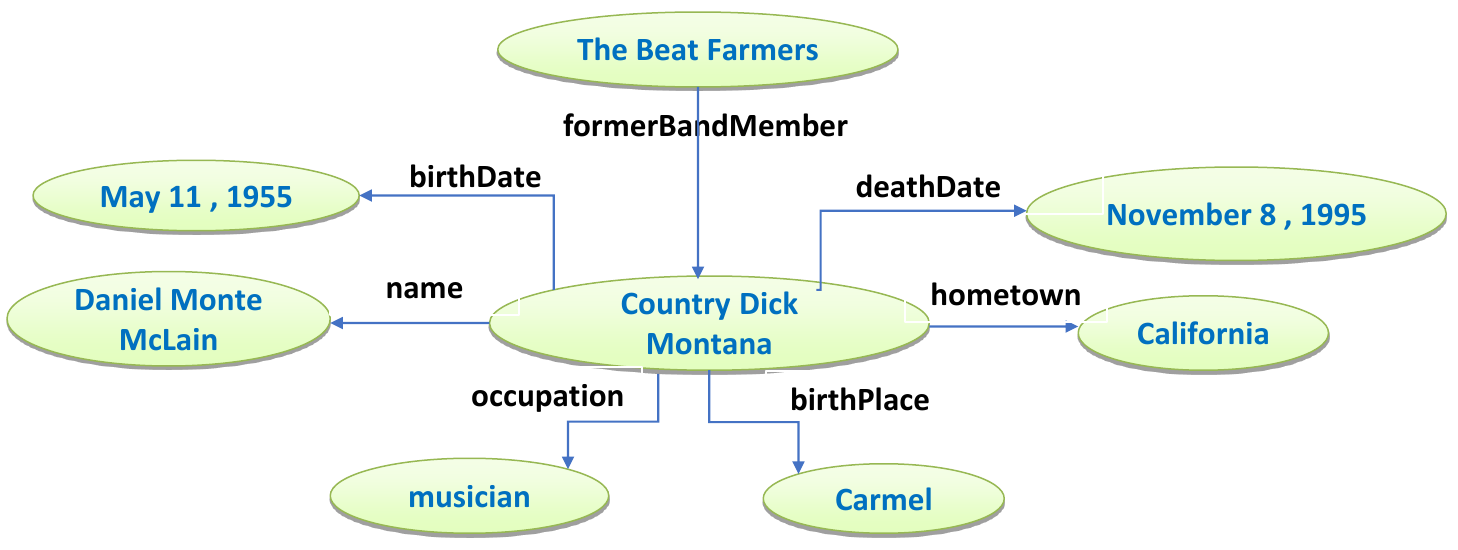}
  \end{center}\\[-0.7em]
 % inserts table %heading
\hline
\\[-0.5em]
\textbf{\underline{Ground-truth text:}} \\\\[-0.8em] Daniel Monte McLain ( May 11 , 1955 – November 8 , 1995 ) , known by the stage name Country Dick Montana , was a musician best known as a member of The Beat Farmers . Montana was born in Carmel , California .
\\ [0.3em]
\hline
\\[-0.5em]
\textbf{\underline{JointGT model:}} \\\\[-0.8em] \textcolor{blue}{Montana was born on May 11 , 1955 in Carmel , California }. \\  [0.3em]
\hline
\\[-0.5em]
\textbf{\underline{CycleGT model:}} \\\\[-0.8em]
\textcolor{blue}{Daniel Monte McLain  (  May 11 , 1955  in  Carmel  ,  Montana  –  November 8 , 1995  in  Carmel  ,  California  ) was a  musician}  , best known as \textcolor{red}{the founder of the band  Country Dick Montana}  .\\  [0.3em]
\hline
\\[-0.5em]
\textbf{\underline{GMP model:}} \\\\[-0.8em]
\textcolor{blue}{Daniel Monte McLain ( May 11, 1955 – November 8, 1995 ), known professionally as Country Dick Montana}, was an \textcolor{blue}{American singer}, \textcolor{red}{songwriter}, and \textcolor{blue}{musician}.\\  [0.3em]
\hline
\\[-0.5em]
\textbf{\underline{Our framework:}} \\\\[-0.8em]
 \textcolor{blue}{Daniel Monte McLain ( May 11 , 1955 – November 8 , 1995 ) was an American musician} . \\ [0.3em]
\hline
\end{tabular}
\caption{Example of generated text on the Genwiki dataset. Here \textcolor{red}{red} colored text represents \textcolor{red}{hallucinated information} and \textcolor{blue}{blue} colored text represents \textcolor{blue}{faithful information}.} % title of Table
\label{fig:genwiki}
\end{figure*}

\section{Comparing Our Result with ChatGPT}
\label{comparison-with-chatgpt}
We randomly take $1000$ sample graphs from the House dataset. Our experiments are conducted using the API of Chat-
GPT (gpt-3.5-turbo) model. We input ChatGPT the sample graphs in a linearized format and asked to summarize the linearized graphs in a real-estate advertising format. We experiment with ChatGPT-ZeroShot (without giving any reference text) , ChatGPT-$k$-FewShot, (where $k$ represents the number of noisy ground-truth text sample is given to ChatGPT as a reference in addition to the input linearized graph) and compare these with our full model. 

Table~\ref{table:chatgpt_results} shows that in terms of faithfulness metrics (BARTScore), ChatGPT-ZeroShot has the best performance. This is because, ChatGPT is a large model and ChatGPT-ZeroShot generates text without taking any noisy ground-truth text as a reference. Whereas, our model is a small (BART-base/T5-base) language model and the model is trained with the full noisy training House dataset. We also notice that the performance of ChatGPT-$k$-FewShot drops with the increase of number of noisy reference text samples. Thus, the more we increase the number of noisy ground-truth texts as a reference to ChatGPT, the more ChatGPT generates hallucinated text similar to ground-truth text. That's why the BLEU, METEOR and ROUGE-L scores increase and BARTscore, FactCC scores decrease with the increase of few shot samples.

We also compare the results using ChatGPT-based evaluation. Table~\ref{table:chatgptgenerations} shows the average of precision, recall and hallucinations which we compute using ChatGPT. The results also show that ChatGPT-ZeroShot performs best in all metrics as usual. Our model outperforms ChatGPT-3-FewShot in terms of precision (higher precision) and hallucination (lower hallucination).

\paragraph{Performance Based on Salient Facts:} We rank in descending order the features (type-wise) of the house graph based on their frequency of occurrence in the House trainining dataset. We take top ten features as \textit{salient} facts. The salient facts are: 1) house\_location, 2) house\_property-type, 3) num. of bedrooms, 4) num. of bathrooms, 5) num of parking spaces, 6) has\_ac, 7) has\_dining, 8) has\_heating, 9) has garage\_spaces and 10) nearest\_train\_station. Using ChatGPT, we enumerate the presence of these facts and measure salient precision, $P_{salient}$ and salient recall, $R_{salient}$ as follows. %based on the following formula~\ref{eqSalient} and formula~\ref{eqSalient2} respectively.

{\small{\begin{align}\label{eqSalient}
  P_{salient} &= \frac{\text{\emph{ \# salient common facts}}}{\text{\emph{\# output facts}}}
\end{align}}
}
{\small{\begin{align}\label{eqSalient2}
  R_{salient} &= \frac{\text{\emph{\# salient common facts}}}{\text{\emph{\# salient input facts}}}
\end{align}}
}
The results from Table:~\ref{table:chatgptsalientfeatures} shows that our model achieves the best average salient precision, $P_{salient}$, and ChatGPT-ZeroShot achieves the best average salient recall. The reason behind this result is that ChatGPT-ZeroShot generated output text contains mostly all the facts from the input graph, whereas our model generated output text gives more focus on the salient facts.
\begin{table}[htbp]
\small
\centering % used for centering table
\begin{tabular}{c|c|c}% centered columns  (2 columns)
\hline %inserts double horizontal lines
\textbf{Generation }  & Avg. & Avg. \\
\textbf{Model}  & Salient Precision & Salient Recall \\[0.5ex]
 % inserts table %heading
\hline
& & \\[-0.7em]
ChatGPT-ZeroShot & 26.75 & \textbf{92.66} \\  [0.2em]
ChatGPT-3-FewShot & 30.27 & 86.36 \\  [0.2em]
\hline
& & \\[-0.7em]
\textbf{Our Full Model} & \textbf{31.64} & 77.16\\ [0.2em]
\hline
\end{tabular}
\caption{ChatGPT Evaluation Results based on $50$ samples from the House dataset considering salient features. \textbf{Bold} fonts denote the best results.} % title of Table
\label{table:chatgptsalientfeatures}
\end{table}

\section{Generated Samples}\label{sec:samples}
Figure~\ref{fig:houseExample} and Figure~\ref{fig:genwiki} show qualitative examples of sample graphs, the ground-truth texts and the texts generated by different models on House dataset and Genwiki dataset, respectively.

\end{document}